\documentclass[
]{ceurart}

\usepackage{listings}
\usepackage{todonotes}
\usepackage{makecell}
\usepackage[capitalize]{cleveref}

\usepackage{pgfplots}
\pgfplotsset{compat=1.18}
\usepackage{pgfplotstable}

\begin{document}

\copyrightyear{2025}
\copyrightclause{Copyright for this paper by its authors.
  Use permitted under Creative Commons License Attribution 4.0
  International (CC BY 4.0).}

\conference{CLEF 2025 Working Notes, 9 -- 12 September 2025, Madrid, Spain}

\title{Multi-Label Plant Species Prediction with Metadata-Enhanced Multi-Head Vision Transformers}


\title[mode=sub]{Notebook for the LifeCLEF Lab at CLEF 2025}


\author[1]{Hanna Herasimchyk}[%
email=hanna.herasimchyk@uni-hamburg.de,
]
\cormark[1]

\author[1]{Robin Labryga}[%
email=robin.labryga@uni-hamburg.de,
]
\cormark[1]

\author[1]{Tomislav Prusina}[%
email=tomislav.prusina@uni-hamburg.de,
]
\cormark[1]

\cortext[1]{These authors contributed equally.}

\address[1]{University of Hamburg,
   177 Mittelweg, Hamburg, 20148, Germany}

\begin{abstract}
We present a multi-head vision transformer approach for multi-label plant species prediction in vegetation plot images, addressing the PlantCLEF 2025 challenge.
The task involves training models on single-species plant images while testing on multi-species quadrat images, creating a drastic domain shift.
Our methodology leverages a pre-trained DINOv2 Vision Transformer Base (ViT-B/14) backbone with multiple classification heads for species, genus, and family prediction, utilizing taxonomic hierarchies.
Key contributions include multi-scale tiling to capture plants at different scales, dynamic threshold optimization based on mean prediction length, and ensemble strategies through bagging and Hydra model architectures.
The approach incorporates various inference techniques including image cropping to remove non-plant artifacts, top-n filtering for prediction constraints, and logit thresholding strategies.
Experiments were conducted on approximately 1.4 million training images covering 7,806 plant species.
Results demonstrate strong performance, making our submission 3rd best on the private leaderboard.
Our code is available at \url{https://github.com/geranium12/plant-clef-2025/tree/v1.0.0}.
\end{abstract}


\maketitle

\section{Introduction}
\label{sec:introduction}
Vegetation plot inventories are essential in ecological research, enabling the sampling and assessment of biodiversity as well as the monitoring of environmental changes. They generate valuable data that supports ecosystem analysis, biodiversity conservation, and evidence-based environmental decision-making. A standard vegetation inventory examines small quadrats that are rectangular frames of about half a square meter placed on the ground to define specific sampling areas. Trained botanists record all plant species found and quantify their presence using metrics such as biomass, ecological scores, or coverage observed in images.

Integrating machine learning methods into this process could drastically enhance efficiency, enabling broader ecological studies with reduced expert involvement. However, developing models capable of identifying multiple plant species among thousands in a single image remains a significant technical challenge.

Having a quadrat image dataset annotated with all present plant species is crucial, yet expensive and challenging to create due to the numerous species in a given area. In contrast, substantial collections of images containing only single plant species already exist, making it much easier to train single-species classification models.

The PlantCLEF 2025 challenge~\cite{plantclef-2025,lifeclef2025,fungiclef2025} seeks to address this gap by evaluating models designed to predict the presence of multiple plant species in high-resolution quadrat images. 
In this competition, models are trained using single-label images of individual plants but are tested on multi-label quadrat images, highlighting the challenge of domain shift between training and test data.

Our main approach utilizes a vision transformer architecture~\cite{dosovitskiy2020image,oquab2023dinov2} equipped with multiple classification heads, enabling the model to simultaneously predict species, genus, and family from a shared feature extraction backbone. This multi-head design effectively integrates taxonomic knowledge and leverages hierarchical relationships, significantly enhancing the robustness of species predictions in complex vegetation plot images.

Key contributions of our work towards improving multi-label classification of plant species in quadrat images include:
\begin{itemize}
    \item We use multi-head predictions and static knowledge of plant taxonomy to harness information contained in the metadata of the training images.
    \item We introduce multi-scale tiling to improve the model's ability to recognize plants at different scales in quadrat images.
    \item We dynamically determine prediction thresholds by optimizing for the mean prediction length.
    \item We utilize bagging to enhance the model's robustness and generalization capabilities.
\end{itemize}
Our code is available on GitHub\footnote{\url{https://github.com/geranium12/plant-clef-2025/tree/v1.0.0}}.

\section{Background}

\subsection{Data}

The training dataset consists of approximately 1.4 million images (about 281 GB) of individual plants, each accompanied by metadata. This large scale presents a significant computational challenge for model training. The dataset, also used in the PlantCLEF 2024 competition, covers 7,806 plant species, 1,446 genera, and 181 families.

\begin{figure}[h!]
    \centering
    \includegraphics[width=0.24\linewidth,height=0.30\linewidth]{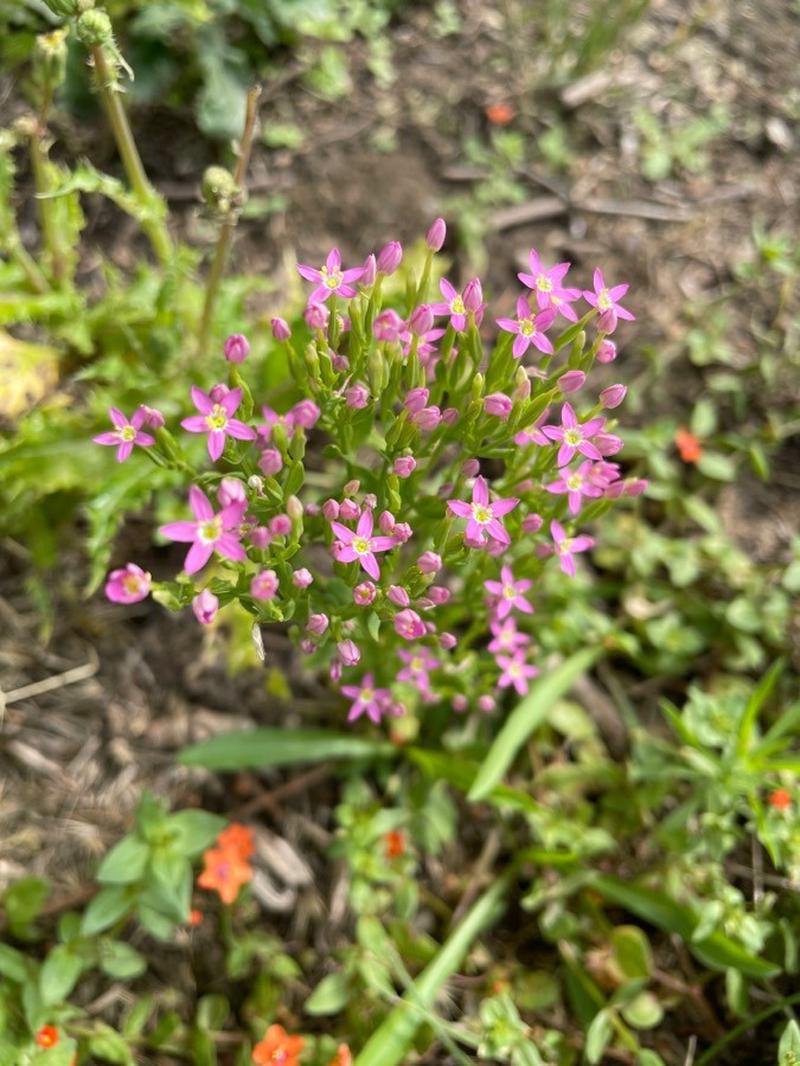}
    \includegraphics[width=0.24\linewidth,height=0.30\linewidth]{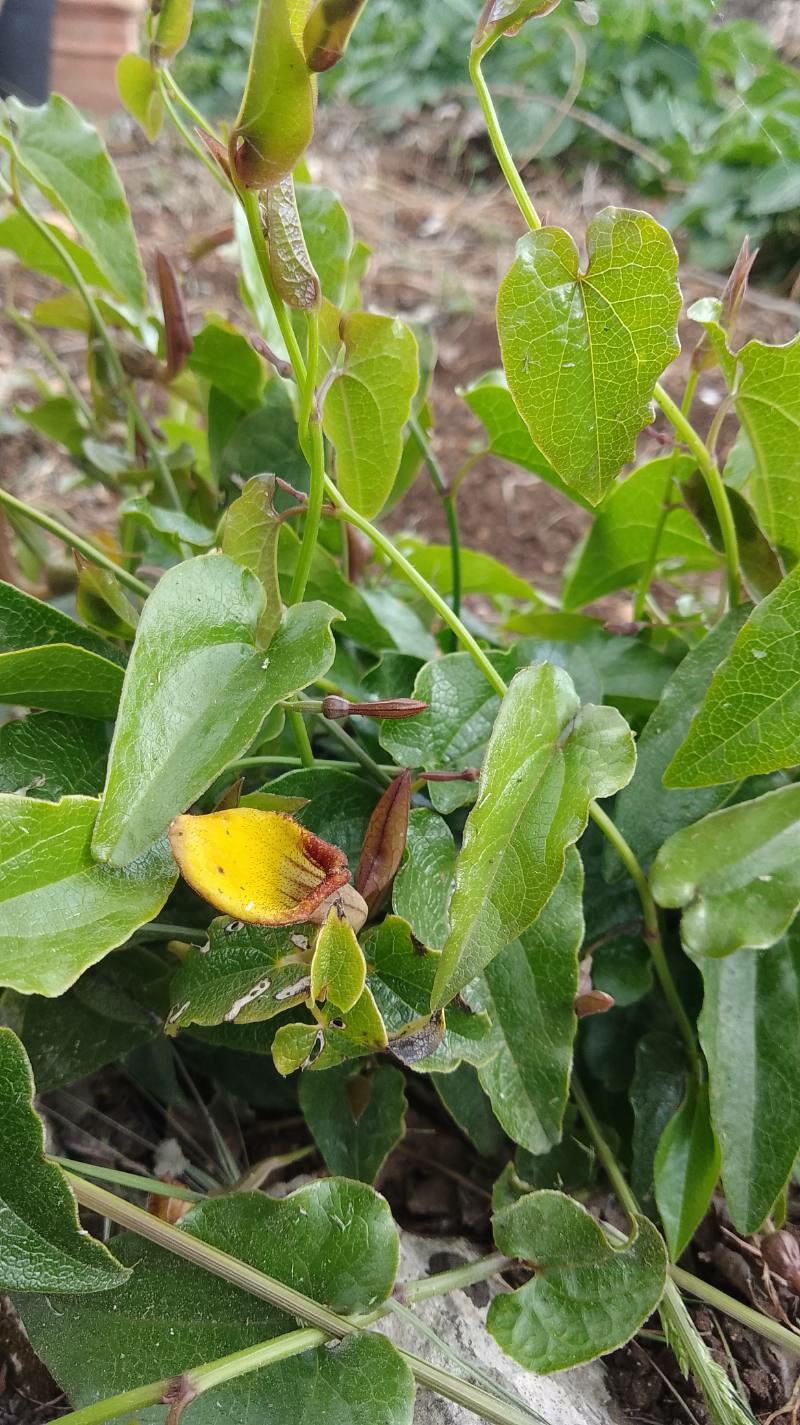}
    \includegraphics[width=0.24\linewidth,height=0.30\linewidth]{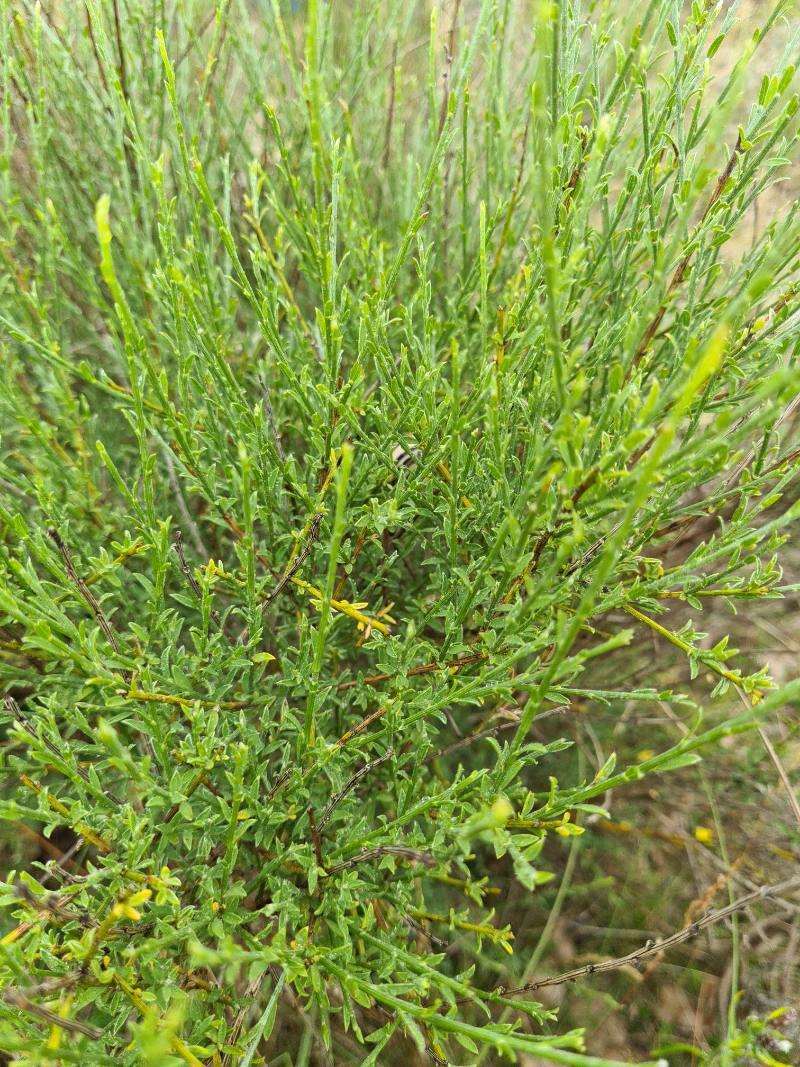}
    \includegraphics[width=0.24\linewidth,height=0.30\linewidth]{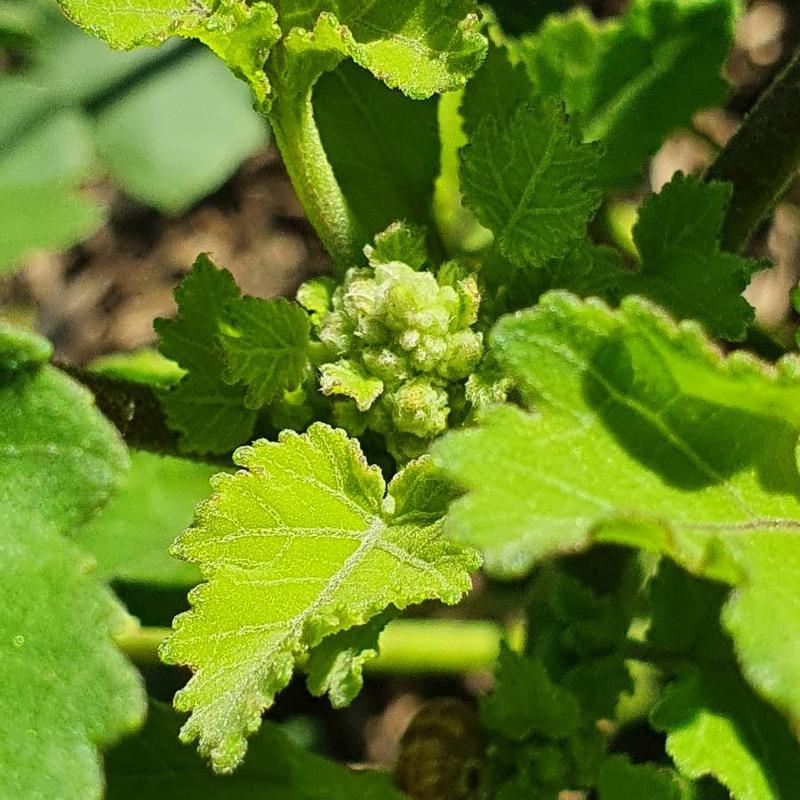}
    \includegraphics[width=0.24\linewidth,height=0.30\linewidth]{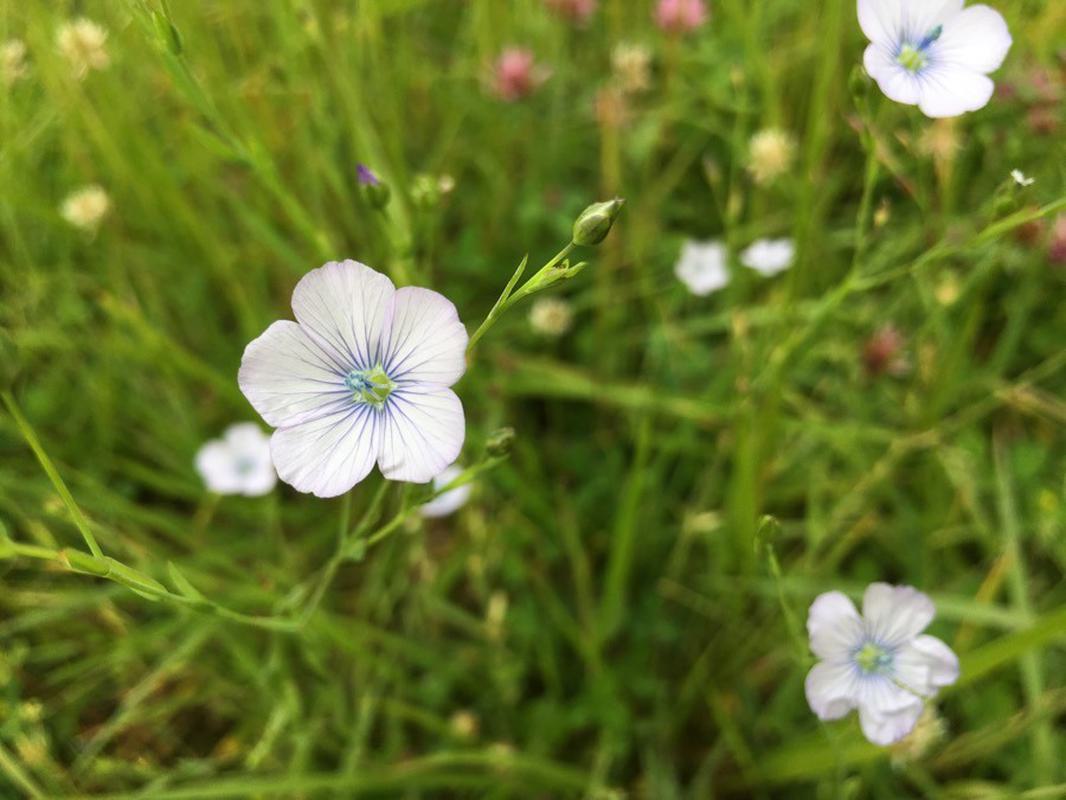}
    \includegraphics[width=0.24\linewidth,height=0.30\linewidth]{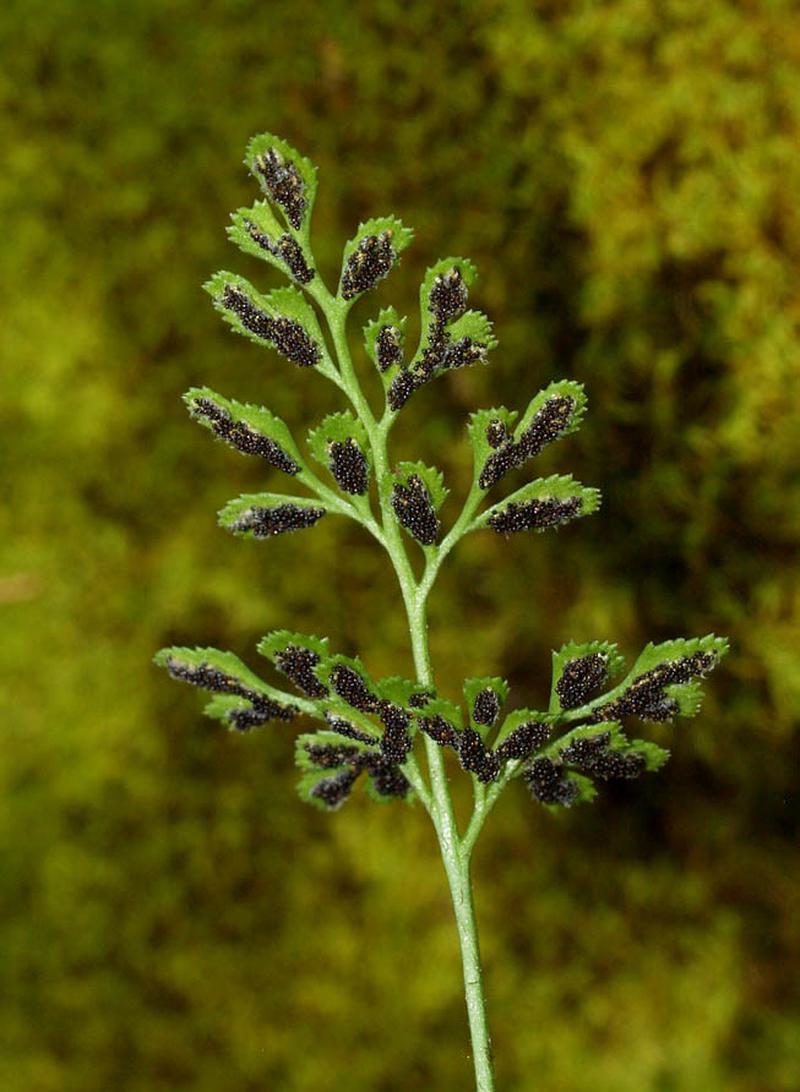}
    \includegraphics[width=0.24\linewidth,height=0.30\linewidth]{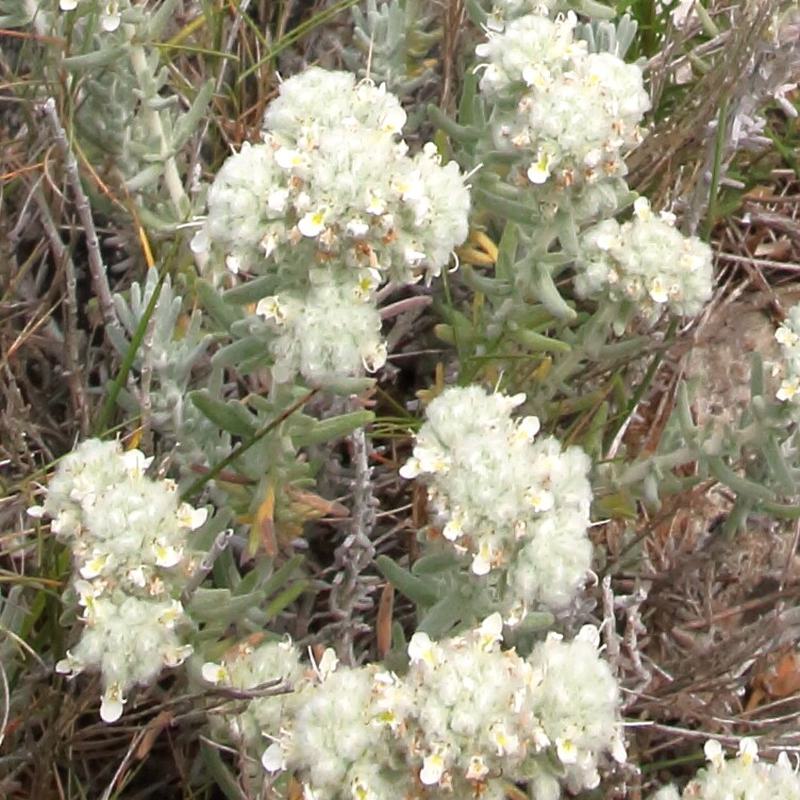}
    \includegraphics[width=0.24\linewidth,height=0.30\linewidth]{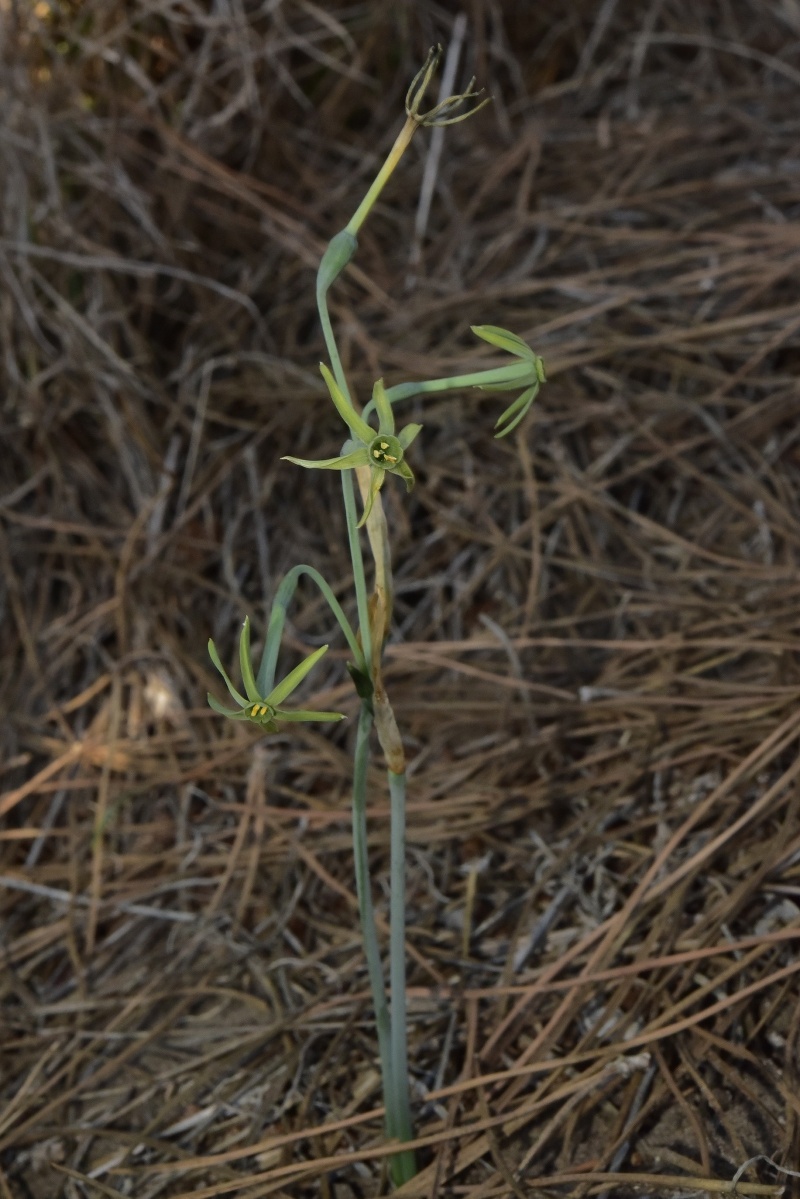}
    \caption{Examples of training images.}
    \label{fig:train-examples}
\end{figure}

The distribution of images across species is shown in \Cref{fig:species-images}, while the distribution of species across genera and families is depicted in \cref{fig:species-distribution}.
Each image is labeled with a single plant species, single genus, and single family, and includes metadata such as organ type and geographic location.
A genus describes a group of plant species, while a family describes a group of plant genera.
Example training images are shown in \cref{fig:train-examples}.
\begin{figure}[h!]
    \centering
    \begin{tikzpicture}
        \begin{axis}[
            width=0.67\linewidth,
            height=0.4\linewidth,
            xlabel={Species rank (sorted by image count)},
            ylabel={Image count},
            grid=major,
        ]
        \addplot [ultra thick, mark=none, color=blue] table [x=species_rank, y=species_counts, col sep=comma] {figures/sspecies.csv};
        \end{axis}
    \end{tikzpicture}
    \caption{Distribution of images per species. Half of all images belong to the 1,787 most common species, while 90\% of images are from the 4,336 most common species, indicating a bias in species representation.}
    \label{fig:species-images}
\end{figure}
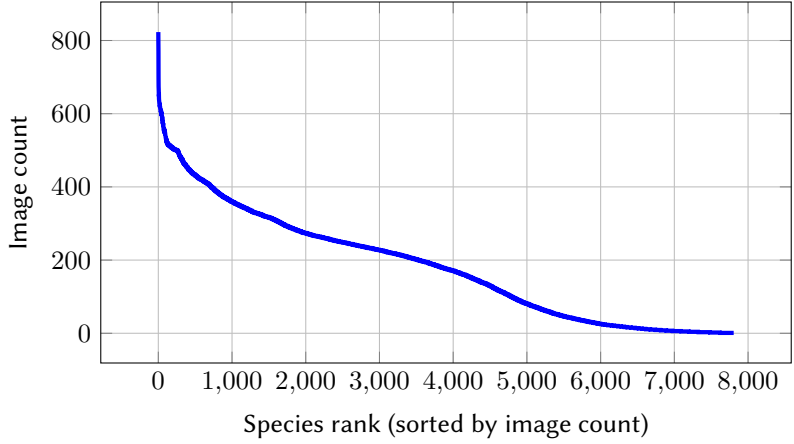

\begin{figure}[h!]
    \centering
    \begin{tikzpicture}
        \begin{axis}[
            width=0.45\linewidth,
            height=0.4\linewidth,
            xlabel={Genus rank (sorted by species count)},
            ylabel={Species count},
            grid=major,
        ]
        \addplot [ultra thick, mark=none, color=blue] table [x=genus_rank, y=genus_counts, col sep=comma] {figures/sgenus.csv};
        \end{axis}
    \end{tikzpicture}
    \begin{tikzpicture}
        \begin{axis}[
            width=0.45\linewidth,
            height=0.4\linewidth,
            xlabel={Family rank (sorted by species count)},
            grid=major,
        ]
        \addplot [ultra thick, mark=none, color=blue] table [x=family_rank, y=family_counts, col sep=comma] {figures/sfamily.csv};
        \end{axis}
    \end{tikzpicture}
    \caption{Distribution of species in genus and family respectively. 50\% of all species are found within the 113 largest genera, and 90\% are contained within the 728 largest genera. Similarly, 50\% of species belong to just 9 of the largest families, while 90\% are included in the 49 largest families. Similarly to \cref{fig:species-images}, these percentages indicate a bias in the distribution of genus and family.}
    \label{fig:species-distribution}
\end{figure}
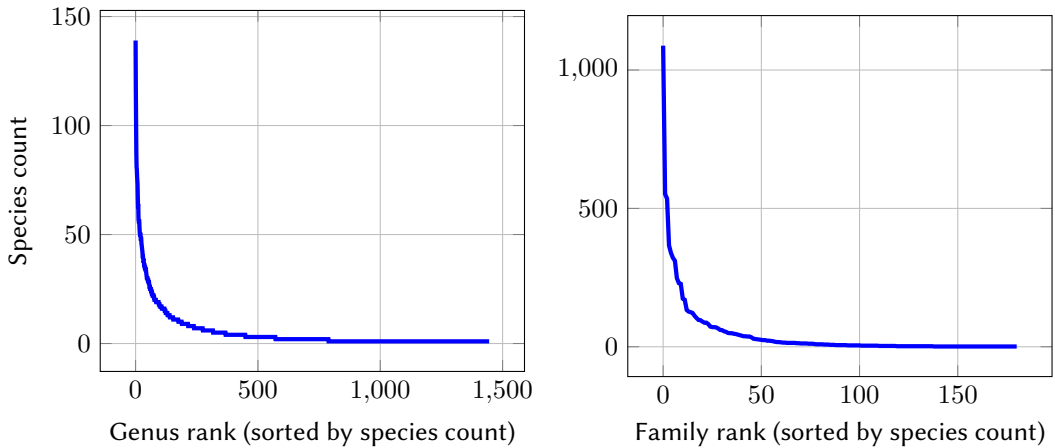

\begin{figure}[h!]
    \centering
    \includegraphics[width=0.33\linewidth,height=0.33\linewidth]{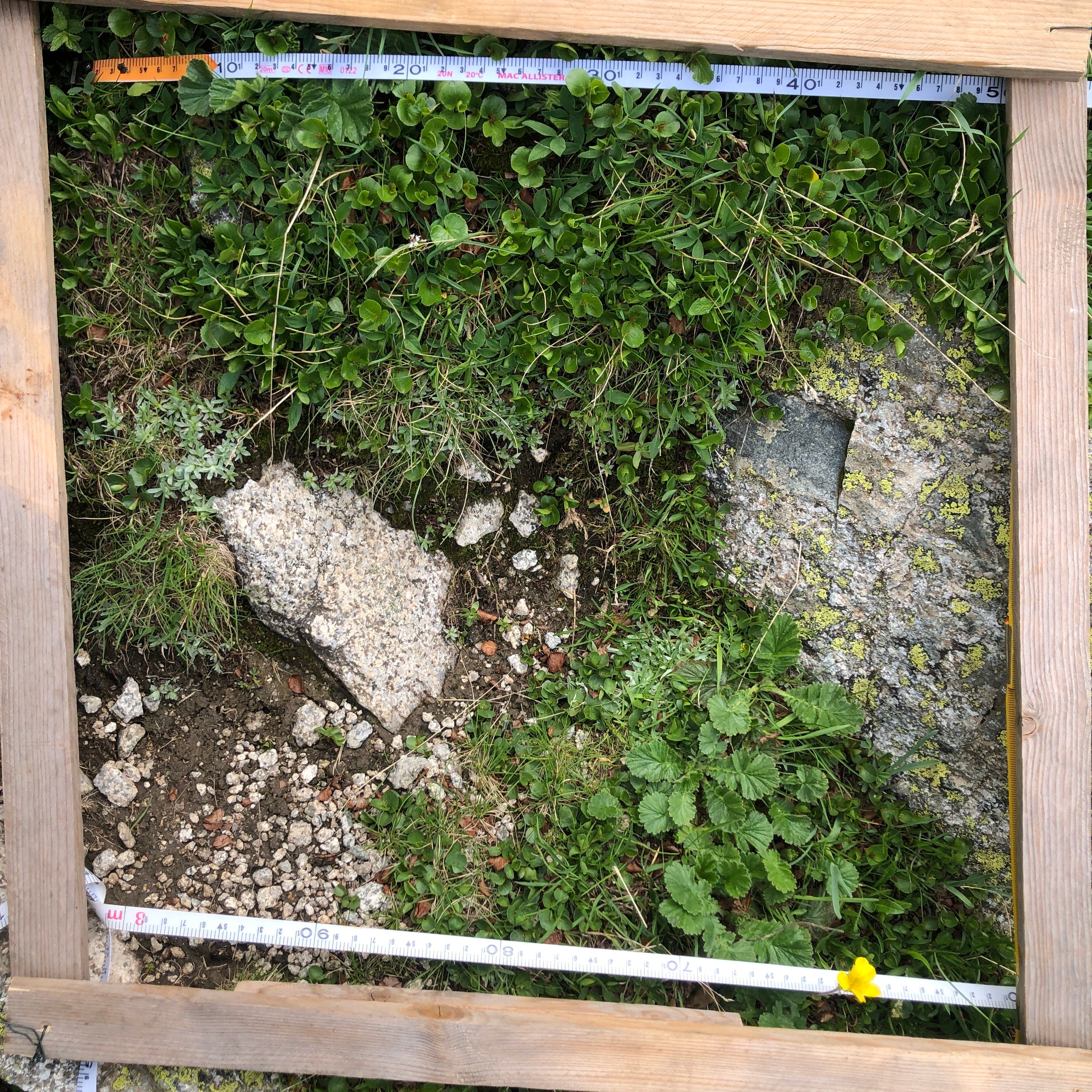}
    \includegraphics[width=0.33\linewidth,height=0.33\linewidth]{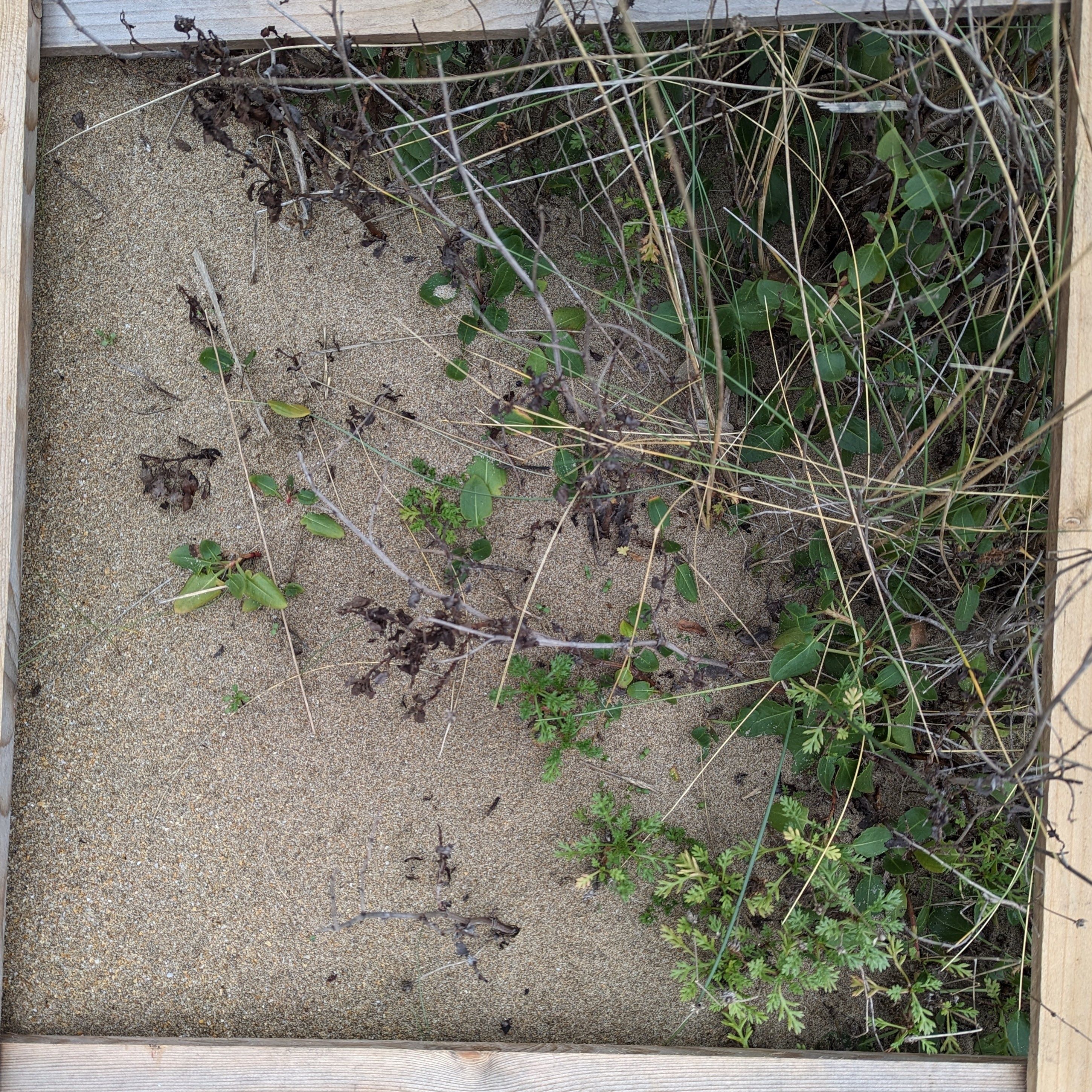}
    \includegraphics[width=0.33\linewidth,height=0.33\linewidth]{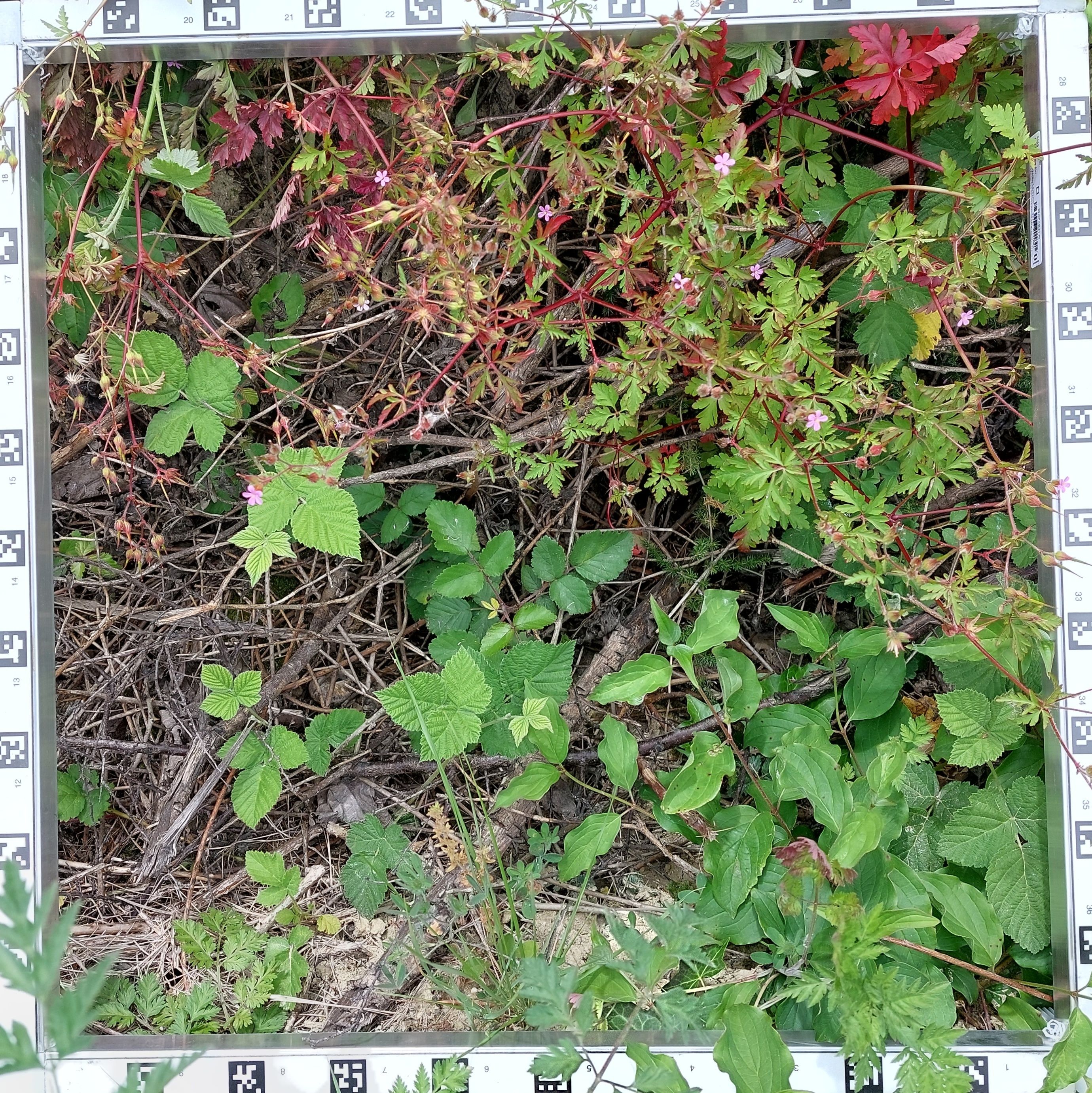}
    \caption{Examples of test vegetation plots.}
    \label{fig:test-examples}
\end{figure}

In contrast, the test dataset comprises vegetative quadrat images containing multiple plant species per image, unlike the single-species focus of the training data. There is no restriction on the number of species present in a test image. Example test images are shown in \cref{fig:test-examples}.

\subsection{Metric}
Unlike PlantCLEF 2024~\cite{plantclef-2024,lifeclef-2024}, this competition uses a modified evaluation metric. The final score is the average of macro-averaged F1 scores, computed for each transect in the test set. A transect is a sequence of vegetation plots (quadrats) placed along a defined path in the field to systematically record species occurrences.

$$
\frac{1}{N} \sum_{i=1}^{N} \left( \frac{1}{T_i} \sum_{j=1}^{T_i} \text{F}1_{ij} \right)
$$

where:
\begin{itemize}
    \item $N$ is the total number of transects,
    \item $T_i$ is the number of quadrats in transect $i$,
    \item $\frac{1}{T_i} \sum_{j=1}^{T_i} \text{F}1_{ij}$ is the macro-averaged F1-score for quadrat $j$ in transect $i$.
\end{itemize}

\subsection{DINOv2 Model}

We used a DINOv2 model~\cite{oquab2023dinov2,goeau10plantclef} provided by the PlantCLEF organizers, pre-trained on single-species training images. The architecture is based on the distilled Vision Transformer Base (ViT-B/14) with registers~\cite{darcet2023vitneedreg} serving as the backbone for feature extraction. For each input image, the model generates an embedding that is then passed through a classification head consisting of one linear layer to predict the species. Further details can be found in~\cite{plantclef-2024}. 

Our choice to use DINOv2 was based on empirical evidence from the PlantCLEF 2024 challenge, where ViT-B architectures demonstrated superior performance compared to alternative model architectures ~\cite{plantclef-2024,gustineli2024multi,foy2024utilizing,chulif2024patchwise}. Furthermore, given the computational constraints associated with the dataset (1.4 million images, 281 GB), training large-scale deep neural networks from scratch would have been computationally prohibitive.
Hence, we used the already pre-trained DINOv2 backbone provided by the organizers without additional finetuning.

\section{Methodology}

\subsection{Training Data Preparation}

For several of our methods, it is necessary to train or retrain models, including the newly added genus and family classifiers, as well as models for distinguishing between plant and non-plant samples. The training procedure we use is described below.

\paragraph{Data Augmentations}
During training, we employed a variety of data augmentation techniques to enhance the model's robustness and generalization capabilities. These augmentations included random cropping, random horizontal and vertical flipping, perspective transformations, and random rotations. Additionally, we applied color jittering to introduce variability in brightness, contrast, and saturation.

We also applied standard normalization and resizing procedure to ensure that input images matched the distribution and size expected by the DINOv2 architecture. This included subtracting the mean and dividing by the standard deviation as well as resizing input images to $518 \times 518$.

\paragraph{Data Split}
The provided training dataset was already pre-split. We decided to use all available data for training, including images that were not original used for pre-training. For internal evaluation, we performed a stratified split of the training data to ensure a balanced representation of species.

\paragraph{LUCAS Dataset}
The organizers provided an additional training dataset called LUCAS (Land Use/Cover Area frame Survey)~\cite{lucas-dataset}, comprising 212,782 unannotated ground vegetation images in a vertical quadrat-like format, amounting to 170GB of data. We explored continued pre-training of the DINOv2 model to incorporate this data, motivated by the idea that exposure to domain-specific vegetation plot imagery during pre-training could enhance the model's representational capacity for downstream classification. However, this approach proved infeasible due to hardware constraints. As a result, we proceeded with the original DINOv2 weights without additional pre-training on the LUCAS dataset.

\subsection{Test Data Preprocessing}

\paragraph{Image Cropping}
Initial visual inspection of the vegetation plot imagery revealed the frequent presence of non-plant artifacts, such as wooden plot frame edges, measuring tapes, and footwear, usually located at the image borders (see \cref{fig:test-examples}). To reduce the influence of these non-plant objects on the model, we experimented with centrally cropping 5\% to 15\% from all four image sides. The 10\% cropping strategy yielded the best results on the public leaderboard, while the 5\% strategy was more effective on the private one, suggesting that the 10\% approach may have been excessive.

\paragraph{Multi-Scale Tiling}
To address the challenge of varying plant sizes and densities within vegetation plots, we implemented a multi-scale tiling approach. This involved splitting the image into a grid of multiple tiles ($2 \times 2$, $3 \times 3$, \dots), allowing the model to capture both small and large plant species effectively. Each tile is used as an input image for the model. All pre-processing steps are applied to each tile accordingly. We additionally experimented with overlapping tiles to ensure that plants on the edges of tiles were not missed. However, we found that using multiple tiles without overlap was sufficient, as the overlap did not lead to any improvement in the results.

\subsection{Model Architecture and Training}

\paragraph{Multi-Head Classification}
To leverage taxonomic information, along the original species MLP classification head, we incorporated additional MLP classification heads for genus and family prediction on top of the DINOv2 ViT-B backbone. These heads utilized metadata associated with each image. We also experimented with the number of layers in each classification head.

Given the strict hierarchical relationship—where each species uniquely belongs to one genus, and each genus to one family, we multiplied the predicted probabilities for species, genus, and family, discarding combinations that do not exist in the provided metadata. This ensured that only valid taxonomic assignments were considered.

In addition to the taxonomic classification heads, we trained a dedicated classification head for organ prediction, designed to identify the type of plant organ depicted in each image (e.g., leaf, flower, stem). However, integrating organ-based information into the overall prediction pipeline proved challenging due to the inherent variability in organ representation among different species.

Furthermore, the dataset included a "scan" organ label indicating images obtained by scanning plants rather than capturing them in natural settings. Since our primary focus is on vegetation plot analysis, which relies on photos of plants in real settings, we hypothesize that removing such images from the training dataset could improve final accuracy.

\paragraph{Hydra Model Architecture}
We used independent classification heads that shared the same embedding from a frozen backbone. Several versions of each head with different numbers of layers were trained simultaneously. During testing, we could swap these pre-trained heads to create various model versions from one main architecture. We refer to this ensemble approach as the Hydra model. The best Hydra model we trained included a one-layer head for species classification and two-layer heads with ReLU activation function in between for genus and family classification.

\paragraph{DINOv2 ViT-L}
We explored scaling the model architecture by training a DINOv2 implementation based on the Vision Transformer Large (ViT-L/14~\cite{dosovitskiy2020image}) backbone. While this architecture offers greater representational capacity compared to smaller variants, preliminary experiments revealed significant computational limitations. A single training iteration on the full PlantCLEF dataset required approximately 30 hours using our GPU cluster (see Model Training in \cref{sec:model-training}). Given the need for at least roughly 50 iterations to achieve convergence, the total training time would exceed 1,500 hours (about 62.5 days), rendering this approach infeasible within the project's resource constraints.

\paragraph{Plant/Non-Plant Filtering}
To reduce false positives from irrelevant foreground clutter (e.g., rocks or soil patches), we trained a binary classifier to distinguish between plant and non-plant regions. We created a separate dataset of non-plant images (primarily rocks) from publicly available sources and trained logistic regression, random forest, and a ViT-based classifier. Out of these three approaches, the Random Forest classifier achieved the highest overall accuracy, correctly identifying plant and non-plant tiles 95\% of the time on our validation data. As a result, we adopted the Random Forest model for filtering of non-plant objects in our primary pipeline. However, the model failed to generalize on the vegetation plot images and did not improve the final prediction quality.

\paragraph{Model Training}
\label{sec:model-training}
We trained the described model architectures on our GPU cluster, utilizing $2 \times$ NVIDIA A6000 GPUs for each experiment. Each ViT-based model was trained for approximately three days, with the duration varying depending on the specific architecture. For detailed technical specifications and code, please refer to our publicly available GitHub repository (see \cref{sec:introduction}).

\subsection{Inference}

We implemented a multi-step prediction pipeline to adapt the single-species classifier to the multi-species quadrat prediction task. Several strategies were empirically tested and integrated, with varying levels of success across the public and private datasets.

\paragraph{Top-n and Bottom-n Filtering}
Given that each vegetation plot image typically contains no more than a dozen distinct plant species, we constrained the number of species predictions per image by limiting the maximum (top-n) allowed predictions. Through experimentation, we found that tuning this upper bound improved scores on the public leaderboard.
The same experiments after the challenge revealed that this often leads to worse performance on the private leaderboard. Additionally, enforcing a minimum (bottom-n) of at least one species prediction per image proved beneficial.

\paragraph{Logit Thresholding}
For each tile, we allowed at most one species contribution to the final prediction. To ensure that only the most confident predictions were included, we applied a logit thresholding strategy.
One approach was to set a minimum logit value for species predictions, filtering out low-confidence predictions. Another approach involved dynamically adjusting the logit threshold based on the mean prediction length across all test images.
To perform this dynamic adjustment efficiently, we utilized pre-computed logits for each test image and tile, and found appropriate thresholds using a bisection search algorithm.
We ended up using dynamically adjusted thresholding with an average of four species per image because of its simplicity of use and apparent performance.

\paragraph{Metadata Merging}
A subset of the test set vegetation plot images included identifiers and dates within their filenames. We investigated whether using image metadata, specifically, merging predictions across images taken in the same field and year, could enhance the score. For example, if a species was identified more than three times across all images of the same plot, it was predicted in every image of that plot. The idea was that such an approach might enhance recall by consolidating information from related plots. However, we did not use this method because: first, this method did not improve our score; second, metadata was not available for the entire test set; third, this method contradicts the goal of the challenge, which is to discover the changes in biodiversity from the vegetation plot.

\paragraph{Bagging}
To further improve the robustness of our predictions, we implemented a bagging strategy (see~\cite{hastie2009elements}). We combined multiple models by averaging their logits from each image tile before generating the final prediction. This method helps reduce variability and increases the reliability of our results by using information from different models.

\paragraph{Kernels}
We implemented a kernel-based smoothing approach applied to the logit outputs of each image tile. Specifically, the logits of neighboring tiles were added to each tile's prediction logits with a weighting coefficient (e.g., 0.5), allowing the predictions of adjacent tiles to influence one another. The idea was that plants might span across tile boundaries. However, initial experiments with kernel-based smoothing did not yield improvements in the final evaluation scores. Consequently, we did not try any alternative kernels. It is likely that the lack of improvement was due to our use of multi-scale tiling, which effectively served a similar purpose.

\paragraph{Other Techniques}
We explored several additional strategies, such as z-score normalization of logits instead of thresholding or filtering out rare species, but observed no consistent improvements across datasets. Due to marginal returns, these methods were ultimately not included in the final pipeline.
\section{Results}

\Cref{tab:leaderboard-top3} shows the top-3 positions in the public and private leaderboards of the PlantCLEF challenge. Our team named "Chlorophyll Crew" achieved the second and third best scores in the public and private leaderboards, respectively.

\begin{table}[h]
\centering
\caption{Top 3 best scores in the public and private leaderboards. Our solution achieved the second and third best scores in the public and private leaderboards respectively under the team name "Chlorophyll Crew".}
\begin{tabular}{llcc}
\toprule
\textbf{Leaderboard} & \textbf{Team name} & \textbf{Rank} & \textbf{Score} \\
\midrule
\multirow{3}{*}{Public} & webmaking & 1 & 0.38132 \\
 & \textbf{Chlorophyll Crew} & 2 & 0.37555 \\
 & TheHeartOfNoise \#Rust \#Candle & 3 & 0.35900 \\
\midrule
\multirow{3}{*}{Private} & TheHeartOfNoise \#Rust \#Candle & 1 & 0.36479 \\
 & DS@GT PlantCLEF & 2 & 0.34489 \\
 & \textbf{Chlorophyll Crew} & 3 & 0.33655 \\
\bottomrule
\end{tabular}
\label{tab:leaderboard-top3}
\end{table}

\Cref{tab:leaderboard-top5} presents our top-5 submissions on both the public and private PlantCLEF leaderboards, as well as our five selected predictions.
While all models achieve higher scores on the public leaderboard, there is a consistent drop in performance on the private leaderboard across all submissions.
This pattern suggests that the public and private test sets are not well-balanced, and that models optimized for the public set may not generalize well to the private set.
The relatively small score differences between submissions on the private leaderboard, contrasted with larger variations on the public leaderboard, further highlight this imbalance.
These results indicate that leaderboard-driven optimization likely led to overfitting on the public test set. In particular, we experienced the smallest drop on the private leaderboard in comparison to the top-performing solutions on the public leaderboard.

\begin{table}[h]
\centering
\caption{
Top-5 submissions on the PlantCLEF leaderboard, showing public and private scores, model configurations, key hyperparameters, and the 5 selected predictions. All submissions use genus-family information, restrict to one species contribution per tile, and require at least one species prediction per image. Bold values indicate top-5 ranking in the respective column. The Hydra model features a single-layer species head along with two-layer heads for genus and family. The 5h1l model, on the other hand, utilizes a single-layer head structure. The Hydra\textsubscript{new} model is an extension of the Hydra model, trained for an additional full epoch. The Vitlarge model represents a ViT-L architecture equipped with two-layer heads. Finally, the pre-trained Vit-B model\protect\footnotemark~was provided by the organizers.
}
\begin{tabular}{lccccccccccc}
\toprule
& logit & \multicolumn{2}{c}{length} & \multicolumn{1}{c}{tiling} & & \multicolumn{2}{c}{score} \\
models & min & mean & max & scales & crop \% & public & private \\
\midrule
\multicolumn{5}{l}{\textbf{Our Top 5 on the Public Leaderboard}} \\
\midrule
\makecell[l]{Hydra\\ 5h1l} & & 4.2 & 9 & 4,5 & 10 & \textbf{0.37555} & 0.33409 \\
\cline{1-1}
\makecell[l]{Hydra\\ Hydra\textsubscript{new}\\ 5h1l\\ 5h1l\textsubscript{new}} & & 4.2 & 9 & 4,5 & 10 & \textbf{0.37543} & 0.33375 \\
\cline{1-1}
\makecell[l]{Hydra\textsubscript{new}\\ 5h1l\textsubscript{new}} & & 4.2 & 9 & 4,5 & 10 & \textbf{0.37542} & 0.33104 \\
\cline{1-1}
\makecell[l]{Hydra\\ 5h1l} & & 4.175 & 9 & 4,5 & 10 & \textbf{0.37540} & 0.33142 \\
\cline{1-1}
\makecell[l]{Hydra\\ 5h1l} & & 4.15 & 9 & 4,5 & 10 & \textbf{0.37522} & 0.33253 \\
\midrule
\multicolumn{5}{l}{\textbf{Our Top 5 on the Private Leaderboard}} \\
\midrule
\makecell[l]{5h1l} & & 4.0 & $\infty$ & 4,5 & 10 & 0.35134 & \textbf{0.34575} \\
\cline{1-1}
\makecell[l]{Hydra\\ 5h1l} & & 4.0 & 9 & 4,5 & 8,10,12 & 0.36263 & \textbf{0.34358} \\
\cline{1-1}
\makecell[l]{5h1l} & 0.01 & & 10 & 4 & 5 & 0.33338 & \textbf{0.34352} \\
\cline{1-1}
\makecell[l]{Hydra\\ 5h1l} & & 4.0 & $\infty$ & 4,5 & 10 & 0.37100 & \textbf{0.34091} \\
\cline{1-1}
\makecell[l]{Hydra\\ 5h1l} & & 3.9 & 9 & 4,5 & 10 & 0.36972 & \textbf{0.34047} \\
\midrule
\multicolumn{5}{l}{\textbf{Our 5 selected submissions}} \\
\midrule
\makecell[l]{5h1l} & 0.02 & & 10 & 4,5 & 10 & 0.36590 & 0.33947 \\
\cline{1-1}
\makecell[l]{Hydra\\ 5h1l\\ Vitlarge} & & 4.2 & 9 & 4,5 & 10 & 0.37305 & 0.33655 \\
\cline{1-1}
\makecell[l]{Hydra\\ 5h1l\\ Vit-B} & & 4.0 & 9 & 4,5 & 10 & 0.36730 & 0.33484 \\
\cline{1-1}
\makecell[l]{Hydra\\ 5h1l} & & 4.2 & 9 & 4,5 & 10 & \textbf{0.37555} & 0.33409 \\
\cline{1-1}
\makecell[l]{5h1l} & 0.02 & & 10 & 1,2,4,5 & 10 & 0.36555 & 0.32074 \\
\bottomrule
\end{tabular}
\label{tab:leaderboard-top5}
\end{table}

Due to a substantial domain shift between the training and test data, we were unable to validate our approaches locally, which forced us to rely on the public leaderboard for model selection.
Despite our efforts to select a diverse set of models, none of our five chosen submissions appear among the top-5 on the private leaderboard, highlighting the challenges presented by the test data split and the limitations of leaderboard-based evaluation.

\footnotetext{\url{https://www.kaggle.com/models/juliostat/dinov2_patch14_reg4_onlyclassifier_then_all/PyTorch/default}}

Our primary multi-head classification approach achieved a substantial improvement over the baseline, which relied on simple single-head plant species classification. As shown in \cref{tab:leaderboard-top5}, all reported results utilize multi-head classification, highlighting this improvement.

We evaluated several hyperparameter configurations and observed that the 10\% cropping strategy yielded the most promising results on the public test set, while the 5\% strategy performed better on the private set, suggesting that the former likely resulted in excessive cropping of informative visual regions.
Top-9 and top-10 filtering did not improve the score, and top-n filtering generally decreased performance on the private leaderboard. Always predicting at least one species positively improved the score.
Dynamically adjusting the threshold with an average of four species per image enhanced the final score.
Our best Hydra model featured a one-layer head for species classification and two-layer heads with an activation function applied between layers for genus and family classification.
Merging metadata did not improve results, likely because metadata was not available for the entire test set and because this approach contradicts the challenge's goal of discovering changes in biodiversity from the vegetation plot.
For multi-scale tiling, we found that using multiple, non-overlapping tiles of sizes 4 and 5 was sufficient, as overlap did not offer any performance gains.
Although plant/non-plant filtering via a Random Forest achieved 95\% validation accuracy on our separate dataset, it failed to generalize to the vegetation plot images and did not enhance the final predictions.
While bagging significantly improved results on the public leaderboard, it had a negative effect on the private leaderboard score. However, bagging did improve the private score when applied to models using different cropping parameters, as seen in our second-best submission on the private leaderboard.
Finally, initial experiments with kernel-based smoothing did not improve the final evaluation scores, possibly because multi-scale tiling already provided a similar effect.

\section{Related Work}

Deep learning and computer vision methods have been widely applied to plant species identification and vegetation analysis. Early work focused on convolutional neural networks (CNNs) for remote sensing and vegetation mapping, as reviewed by Kattenborn et al.~\cite{kattenborn2021cnn}. More recently, transformer-based architectures have shown promise for plant-related tasks, such as weed detection in UAV imagery~\cite{reedha2022transformer}, and our work builds on this trend by utilizing a vision transformer backbone for multi-label plant species prediction.

Patch-based and multi-scale approaches have been explored to address the challenge of varying object sizes in images. Adelson et al.~\cite{adelson1984pyramid} introduced image pyramid methods, which similar to our use of multi-scale tiling captures information at different spatial resolutions.

Hierarchical classification, which exploits for example taxonomic relationships, has been studied in various domains. Silla and Freitas~\cite{silla2011hierarchical} provide a comprehensive survey of hierarchical machine learning. An example of hierarchical classification in the context of taxonomy is the work by Colonna et al.~\cite{colonna2018comparison} that used a top-down approach to predict family, genus, and species in frogs. Several works~\cite{hernandez2013hybrid,fiaschi2024informed} propose multiplying probabilities along the taxonomic hierarchy with some using one classifier per hierarchical layer, and some using one per inner node in the hierarchy. It is similar to our multi-head architecture that predicts species, genus, and family independently and fuses their outputs.

Data augmentation remains a key technique for improving model robustness. Shorten and Khoshgoftaar~\cite{shorten2019survey} provide a comprehensive survey of image augmentation methods, many of which we use in our training pipeline.

Previous work in the PlantCLEF2024 challenge~\cite{plantclef-2024,lifeclef-2024} featured diverse deep learning approaches for plant species identification. Foy and McLoughlin~\cite{foy2024utilizing} leveraged the vision transformer (ViT) architecture together with the Segment-Anything Model (SAM) to effectively suppress false positives in non-plant image regions. Gustineli et al.~\cite{gustineli2024multi} explored multiple embedding methods and classifier architectures based on ViT, while Chulif et al.~\cite{chulif2024patchwise} combined CNNs and ViT with Bayesian Model Averaging for enhanced prediction. These approaches highlight a trend toward vision transformers and advanced post-processing techniques for robust plant species identification.

\section{Conclusion}

We present a metadata-enhanced multi-head vision transformer for multi-label plant species prediction, combining species, genus, and family outputs through taxonomic fusion. Using multi-scale tiling, dynamic thresholding, and ensemble strategies (Hydra), our model achieved strong results on the public leaderboard.

However, performance dropped on the private test set, revealing sensitivity to domain shift and the limitations of leaderboard-based tuning, but still having competitive results. 

Future work should address domain adaptation, incorporate organ-specific cues, and explore fine-tuning strategies to improve real-world robustness.
\begin{acknowledgments}
  We want to thank the organizers of PlantCLEF 2025 and LifeCLEF 2025 for hosting the competition.
\end{acknowledgments}

\section*{Declaration on Generative AI}
During the preparation of this work, the author(s) used ChatGPT, GitHub Copilot, Grammarly in order to: Grammar and spelling check, Paraphrase and reword. After using this tool/service, the author(s) reviewed and edited the content as needed and take(s) full responsibility for the publication’s content.

\bibliography{bibliography}

\end{document}